\documentclass[10pt,twocolumn,letterpaper]{article}

\usepackage{cvpr}      

\usepackage{amsmath}        
\usepackage{enumitem}       

\usepackage{bm}             
\usepackage{booktabs}       
\usepackage{mathtools}      
\usepackage[T1]{fontenc}    
\usepackage[utf8]{inputenc} 
\usepackage{microtype}      

\usepackage{algorithm}
\usepackage{algpseudocode}
\usepackage{booktabs}
\usepackage{tabularx}
\usepackage{arydshln}
\usepackage{graphicx}
\usepackage{caption}   
\usepackage{float}     
\usepackage{makecell} 
\definecolor{myviolet}{HTML}{6666FF}

\definecolor{cvprblue}{rgb}{0.21,0.49,0.74}
\usepackage[pagebackref,breaklinks,colorlinks,allcolors=cvprblue]{hyperref}

\def\paperID{8768} 
\def\confName{CVPR}
\def\confYear{2026}

\title{LINGUAL: Language-INtegrated GUidance in Active Learning for Medical Image Segmentation}

\author{
Md Shazid Islam$^{1}$,
Shreyangshu Bera$^{1}$,
Sudipta Paul$^{2}$,
Amit K. Roy-Chowdhury$^{1}$\\
$^{1}$UC Riverside\quad
$^{2}$Samsung Research America\\
{\tt\small misla048@ucr.edu, sbera004@ucr.edu, spaul007@ucr.edu, amitrc@ucr.edu}
}

\begin{document}
\maketitle
\begin{abstract}






Although active learning (AL) in segmentation tasks enables experts to annotate selected regions of interest (ROIs) instead of entire images, it remains highly challenging, labor-intensive, and cognitively demanding due to the blurry and ambiguous boundaries commonly observed in medical images. Also, in conventional AL, annotation effort is a function of the ROI: larger regions make the task cognitively easier but incur higher annotation costs, whereas smaller regions demand finer precision and more attention from the expert. In this context, language guidance provides an effective alternative, requiring minimal expert effort while bypassing the cognitively demanding task of precise boundary delineation in segmentation. Towards this goal, we introduce $\textbf{LINGUAL}$: \underline{L}anguage-\underline{IN}tegrated \underline{GU}idance in \underline{A}ctive \underline{L}earning — a framework that receives natural language instructions from an expert, translates them into executable programs through in-context learning, and automatically performs the corresponding sequence of sub-tasks without any human intervention. We demonstrate the effectiveness of \textit{LINGUAL} in active domain adaptation (ADA) achieving comparable or superior performance to AL baselines while reducing estimated annotation time by approximately $80 \%$.

\end{abstract}


\section{Introduction}

Deep learning has revolutionized medical image segmentation using large volumes of annotated data \citep{van2014transfer}. However, scaling up pixel-level annotation of medical images is time-consuming and labor-intensive. To alleviate this annotation burden, active learning (AL) \cite{islam2025odes,xie2022towards,luo2024uncertainty} has emerged as a promising paradigm. The active learning process iteratively identifies the most informative regions to annotate for effective model training. As a result, we can achieve competitive segmentation performance while substantially reducing the manual labeling effort.

\begin{figure}
    
    \centering
    
    \includegraphics[width=0.9\linewidth]{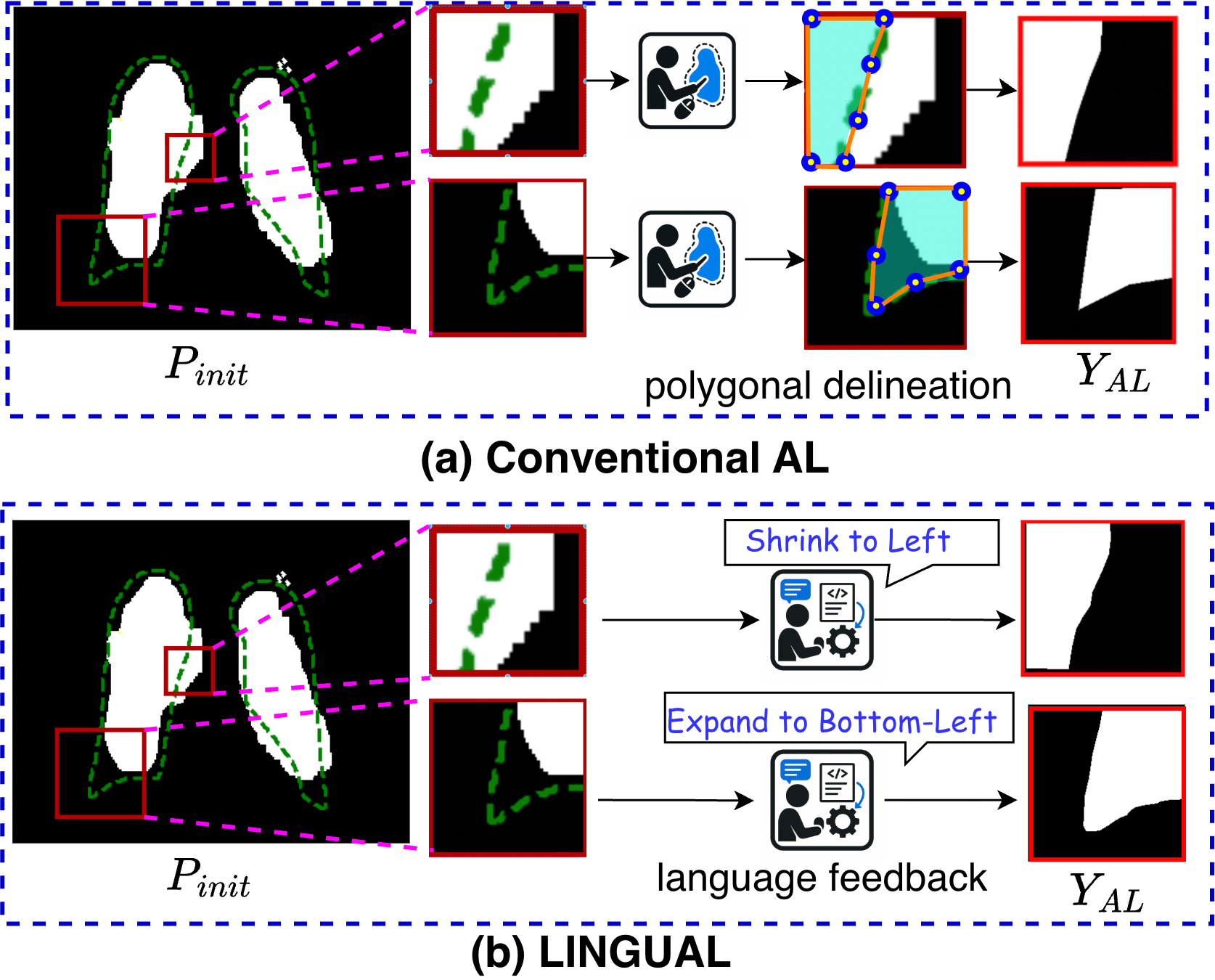}
    \caption{In both (a) and (b), $P_{init}$ denotes the initial prediction map, where two regions of interest (ROIs) are selected for expert review. Green dotted line indicates ground truth boundary (a) In conventional AL, the expert manually annotates the ROIs through polygonal delineation to obtain refined segmentation $Y_{AL}$. (b) In contrast, \textit{LINGUAL} enables the expert to provide high-level language feedback which is executed autonomously to achieve equivalent corrections as in (a) , thereby reducing manual effort. }

    \label{fig:introduction}
\end{figure}

Although active learning focuses on selection and refinement of informative sub-regions of an image, it still requires dense pixel-level annotation of those selected sub-regions. Refining those sub-regions in medical image segmentation requires expert knowledge and is often challenging due to ambiguous or blurry boundaries \cite{lee2020structure,yuan2024lcseg}. At the same time, it requires substantial cognitive effort and precision from experts during correction through manual annotation. So we ask ourselves: \textit{"Can we efficiently refine the segmentation maps of medical images within an active learning framework, achieving high precision and utilizing minimal human effort?"}


 In this context, natural language based feedback offers a compelling alternative—allowing experts to express high-level intent through language rather than performing exhaustive pixel-level annotations. For example, issuing natural commands such as “remove the small fragments” or “expand the boundary of the left” enables experts to convey high-level semantic intent to guide segmentation refinement bypassing tedious annotation process. Motivated by this observation, we propose \textbf{\textit{LINGUAL}}, \underline{L}anguage-\underline{IN}tegrated \underline{GU}idance in \underline{A}ctive \underline{L}earning, to the best of our knowledge, is the first to incorporate language feedback into the AL segmentation refinement process. Figure \ref{fig:introduction} exhibits the comparison between conventional AL and our proposed method \textbf{\textit{LINGUAL}}.


    
    
    


\begin{figure}
    
    \centering
    
    \includegraphics[width=0.9\linewidth]{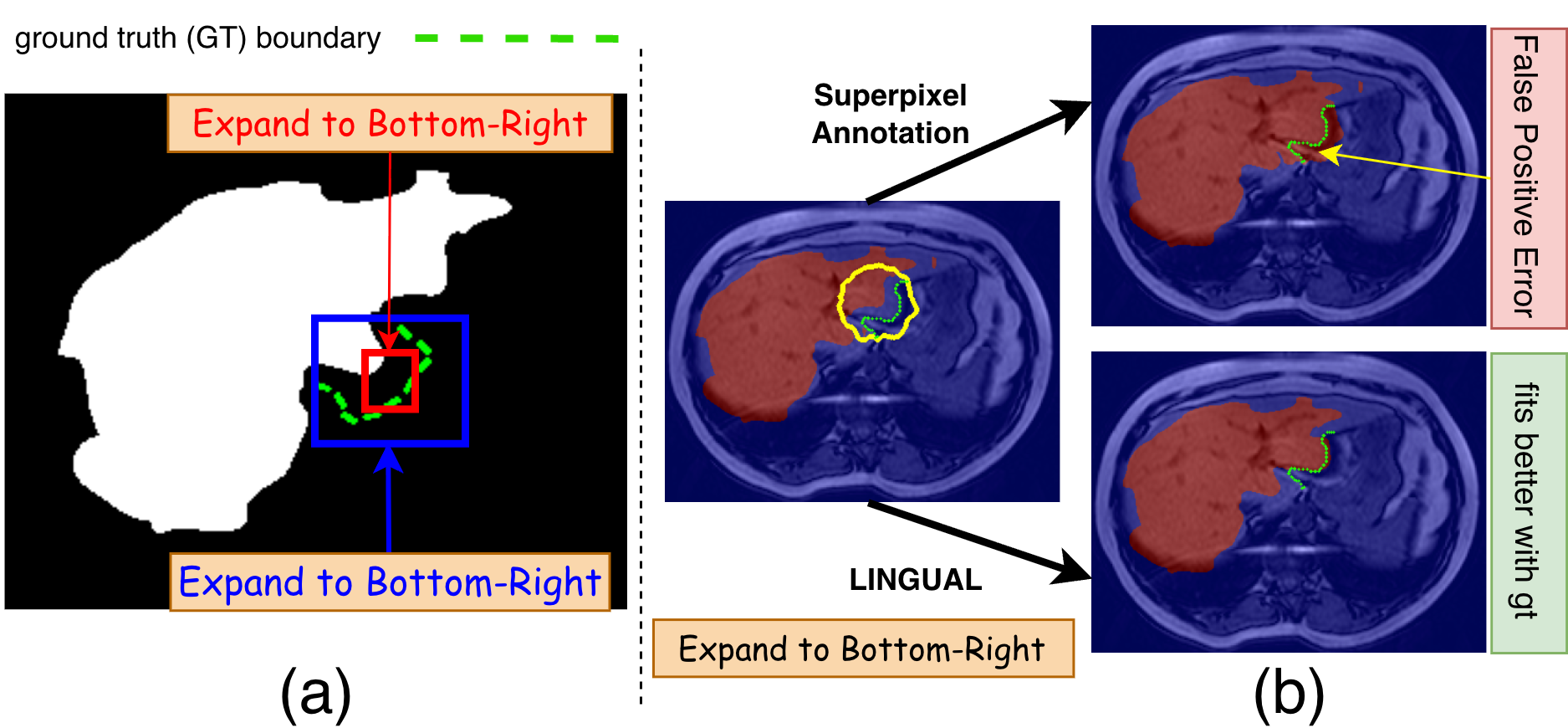}
    
    \caption{(a) On the initial prediction map, 2 ROIs have been chosen (Red box - smaller, Blue box -larger).     
    Conventional AL methods would need higher manual effort to annotate the Blue ROI. However, \textit{LINGUAL} enables same language instruction ("Expand to Bottom-Right") to perform required corrective operation irrespective of ROI size, keeping human effort unchanged. (b) Area inside yellow contour indicates a super-pixel containing regions both from inside and outside of ground truth (GT) boundary. Annotating this entire super-pixel as foreground leads to false positive error. In contrast, Executing ``Expand to Bottom-Right" command through \textit{LINGUAL} reduces annotation error by fitting updated boundary more precisely with the GT boundary.}
    \label{fig:advantage_LINGUAL}
\end{figure}



Labeling for segmentation is generally performed through software-driven delineation, where experts outline target regions by placing sequential polygonal vertices along object boundaries~\cite{berg2019ilastik,russell2008labelme}. This process is challenging due to the irregular and complex boundaries which are common in medical images. In contrast, the proposed \textit{LINGUAL} framework decomposes the annotation process into two complementary stages: (i) language feedback from the expert, (ii) autonomous  planning and execution. The expert provides high-level natural language instructions, which are translated into executable programs through in-context learning \cite{gupta2023visual}. These programs define a sequence of operations—such as expanding, shrinking, or smoothing boundaries—that are then executed automatically to refine the segmentation.

Existing AL methods follow annotation budget constraints which often limits the scalability of the process. When larger regions of interest (ROI) are selected, annotation costs escalate quickly. Conversely smaller patches may reduce annotation cost, but they are very inconvenient for experts to annotate and often lack the necessary contextual information. In contrast, \textit{LINGUAL} maintains nearly constant manual effort from the expert, even as the ROI size increases (Figure \ref{fig:advantage_LINGUAL}a). For a given region, language commands such as “remove small fragments at the bottom,” “shrink to the left,” or “fill up the holes” remain consistent regardless of the ROI’s scale within a reasonable limit. The automation cost may increase with larger ROI, but the human level effort (language instruction) remains almost unchanged.  Thus \textit{LINGUAL} enhances the scalability of AL without increasing human effort.

\begin{figure*}[htbp]
    \centering
    
    \includegraphics[width=\textwidth]{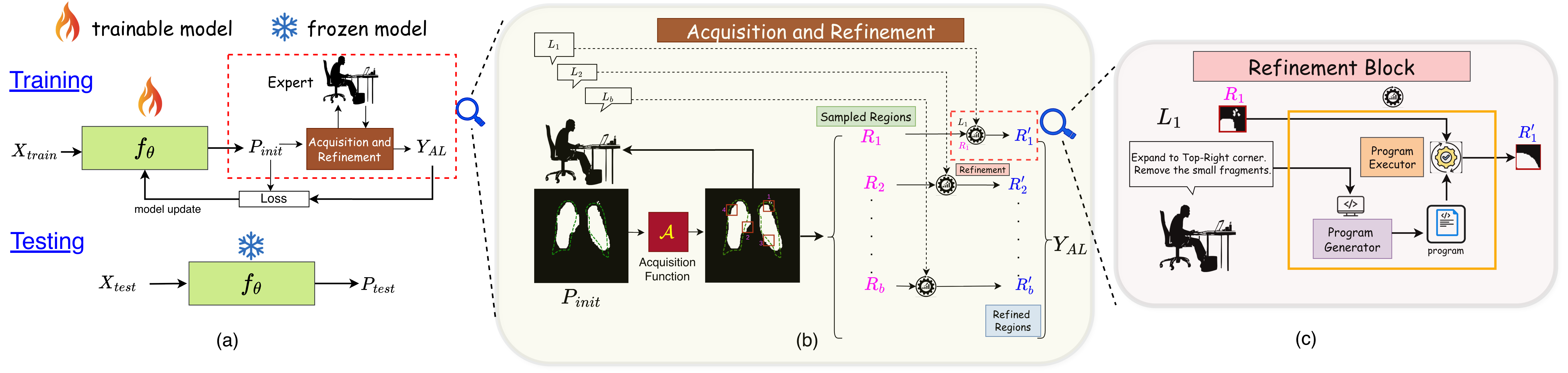}  

    \caption{ \textbf{Workflow Overview:} (Figure 2a)  A pretrained source model $f_{\theta}$ is adapted to the target domain through active domain adaptation (ADA). Given the training data in target domain $X_{\text{train}}$, $f_{\theta}$ produces an initial prediction map $P_{\text{init}}$. Through acquisition (Figure 2b) a budget number of regions \( \{R_i\}_{i=1}^b \) are sampled where the expert provides language feedback. Each region \( R_i \) is refined using a refinement block (Figure~2c), composed of a Program Generator and Program Executor. The Program Generator translates \( L_i \) into a program (sequence of operations), which the Program Executor applies to produce the refined patch \( R'_i \). Substituting \( \{R'_i\}_{i=1}^b \) for \( \{R_i\}_{i=1}^b \) yields the refined segmentation map \( Y_{AL} \). Subsequently, $f_{\theta}$ is updated using the loss computed between $P_{\text{init}}$ and $Y_{\text{AL}}$. 
 After convergence, the adapted and frozen $f_{\theta}$ is deployed for inference.}

    \label{fig:workflow}

\end{figure*}

There have been some recent works on super-pixel based AL \cite{gao2024efficient, kim2023adaptive, hwang2023active} where the ROI are sampled in form of super-pixels. These super-pixels are generated by automated algorithms \cite{achanta2012slic}, allowing experts to assign only the class labels to them—thus bypassing the need to draw detailed boundaries manually.
However, this simplification becomes problematic in medical images, where boundaries are often blurry or ambiguous \cite{lee2020structure,yuan2024lcseg}. A super-pixel may extend beyond the true boundary, and assigning it a single class label introduces labeling error (Figure \ref{fig:advantage_LINGUAL}b). Training the model with such erroneous labels can degrade its performance. In contrast, \textit{LINGUAL} addresses this issue by interpreting expert language instructions (e.g., “expand to right,” “shrink at top”) as spatial actions within the ROI, iteratively refining the segmentation map through small corrective adjustments until it aligns with the true boundary, thereby minimizing annotation errors.

The main contributions of this work are as follows.
\begin{itemize}[leftmargin=*,noitemsep,topsep=0pt,parsep=0pt,partopsep=0pt]


    \item We introduce \textit{\textbf{LINGUAL}}, the first language-guided AL framework that enables experts to refine segmentation through natural language instructions, effectively reducing reliance on labor-intensive manual delineation based annotation.
    
    \item We design novel modules that translate language instructions into executable programs for adaptive boundary refinement, autonomously expanding or contracting regions to align with expert intent and improve segmentation accuracy.


    

    \item We conduct extensive comparisons against patch-based and superpixel-based active learning baselines under identical annotation budgets, showing that \textit{LINGUAL} attains comparable or superior segmentation performance while reducing the estimated annotation time by approximately 80$\%$

\end{itemize}


\section{Related Works}
\noindent \textbf{\textit{Active Learning (AL) in Segmentation.}} AL reduces annotation effort in segmentation by allowing experts to label only the most informative regions (ROIs), which helps maintain acceptable segmentation accuracy while significantly minimizing manual intervention. These ROIs are selected based on uncertainty \citep{holub2008entropy,luo2024uncertainty}, impurity \cite{xie2022towards,franco2024hyperbolic} or diversity \citep{sener2018active,parvaneh2022active}. There are some hybrid methods such as BADGE \citep{ash2019deep}, ODES \cite{islam2025odes} which integrates uncertainty and diversity based on gradient, spatial and feature based distance. Early AL methods \cite{yang2017suggestive,wang2018interactive} trained models from scratch, whereas recent approaches \cite{islam2025odes,xie2022towards,luo2024uncertainty} leverage pretrained models with strong visual priors, achieving faster convergence and improved sample efficiency.

\noindent \textbf{\textit{Annotation Strategy.}} 
AL annotations are mainly categorized into patch-level \cite{islam2025odes, xie2022towards} and superpixel (SP)-level \cite{hwang2023active,kim2023adaptive,gao2024efficient,cai2021revisiting} approaches. In patch-level AL, sampled regions by $\mathcal{A}$ may span multiple semantic classes, requiring experts to manually delineate boundaries and assign labels accordingly. Several open-source software tools can be employed to refine these regions through manual annotation \cite{russell2008labelme, berg2019ilastik,yushkevich2016itk}. These software tools support polygonal and brush-based annotations, allowing users to manually delineate object boundaries—a process known as manual or polygonal delineation, where experts outline the target regions by placing sequential polygonal vertices along the boundary. In contrast, SP-level AL methods  assume that each sampled SP is semantically homogeneous, allowing experts to assign labels directly without boundary delineation. Here ROIs are local coherent pixel groups which are obtained by automated algorithms like SLIC \cite{achanta2012slic}, SEEDS \cite{van2015seeds}, ERS \cite{ERS}. SP enables experts to assign semantic labels of ROIs by selecting one class from multiple options through a click-based interactive user interface. However, these approaches are susceptible to labeling errors when a superpixel spans multiple instances due to ambiguous or indistinct boundaries \cite{li2020superpixel}.



Due to the high precision required in medical imaging, manual delineation remains the predominant approach for annotation \cite{wang2021annotation}, despite the challenges posed by the irregular and highly complex boundary structures commonly found in medical images. To address these challenges, we propose \textit{LINGUAL}, a simple yet effective framework that introduces automation into the annotation process, thereby substantially reducing direct human involvement while maintaining accuracy.

\section{Methodology}
\label{sec:methodology}
\begin{figure*}[htbp]
    \centering
    \includegraphics[width=0.93\textwidth]{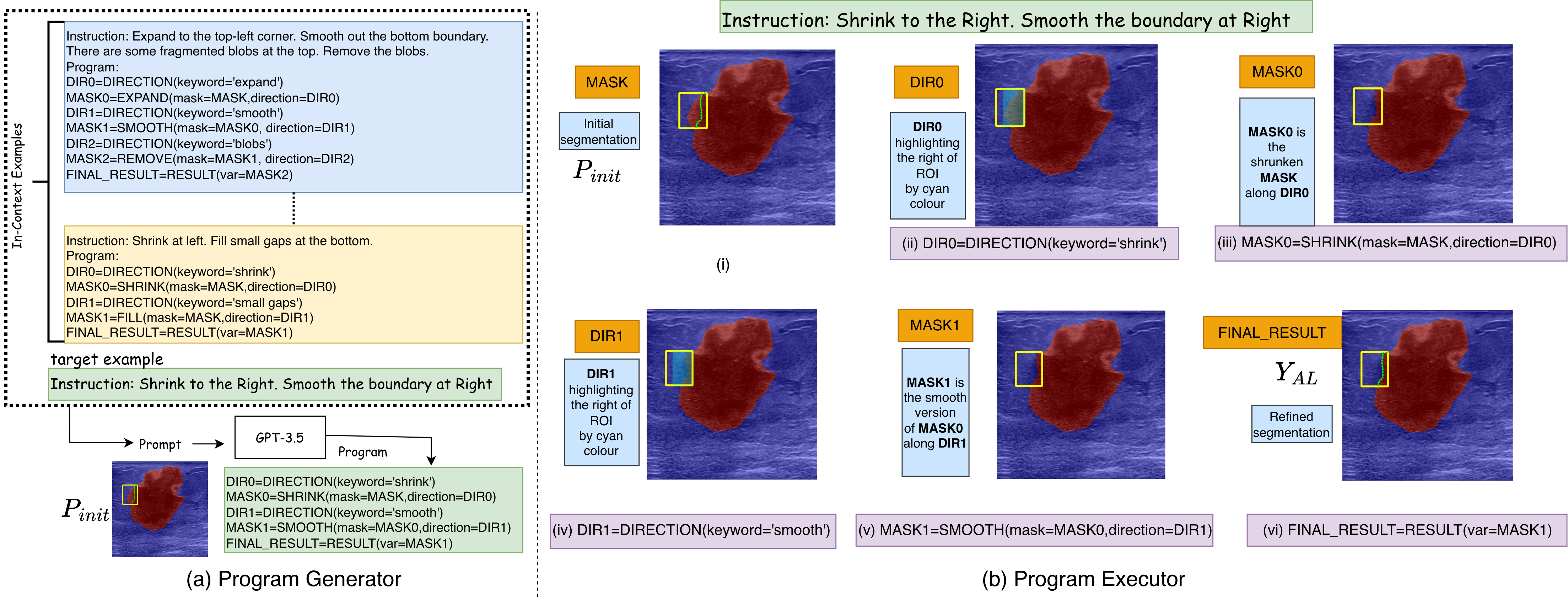}
    \caption{ \textbf{Segmentation refinement by LINGUAL}: (a) \textit{Program Generator }: We show initial prediction map $P_{init}$ (red overlay on input image) along with the ROI (yellow box) and the Ground Truth (GT) boundary (green dotted line) inside ROI . The GT boundary is shown only for illustration, the expert does not use that. In order to refine segmentation, the expert provides natural language instructions (target example) which are translated into program sequences using in-context learning capability of GPT-3.5 with manually crafted in-context examples. (b) \textit{Program Executor}: Each step of generated program (purple text box) is shown along with its output image and output variable name (orange text box), and its interpretation (blue text box). We note that the input image is used during both the EXPAND and SHRINK operations. For simplicity, it is not passed explicitly as a function argument, since the program already has access to it internally.}
    \label{fig:visprog}
\end{figure*}

We demonstrate the application of \textit{LINGUAL} within a domain adaptation framework \cite{huo2022domain,ding2022source}, where active learning is employed to enhance adaptation efficiency—a setting commonly known as Active Domain Adaptation (ADA) \cite{islam2025odes,xie2022towards,luo2024uncertainty} for medical image segmentation. Initially, a segmentation model $f_\theta$ is trained on a set of labeled source data $\mathcal{S} = \{(X_S^i, Y_S^i)\}_{i=1}^{M_S} \sim \mathcal{D}_S$ to segment a total of $C$ number of classes, where $\mathcal{D}_S$ is the source domain data distribution. During adaptation, we are given an unlabeled target domain $\mathcal{T} = \{ X_T^j \}_{j=1}^{M_T} \sim \mathcal{D}_T$, whose distribution differs from $\mathcal{D}_S$ due to domain shift. The objective is to adapt $f_{\theta}$ to perform well on $\mathcal{D}_T$ using acquisition guided language-based feedback. Figure \ref{fig:workflow} demonstrates the overall workflow.
The adaptation process proceeds for a total of $R$ active rounds under a per-image annotation budget of $B\%$. 
In each round, approximately $B\%/R$ of the image area is selected for expert interaction. 
The segmentation model $f_{\theta}$ is adapted on the training data of target domain $\{ X_{\text{train}} \}_{T}$, denoted simply as $X_{\text{train}}$ for brevity. 
Starting from the initial prediction map $P_{\text{init}}$ by $f_{\theta}$, an uncertainty-based acquisition function $\mathcal{A}$ identifies the most informative regions according to the round’s budget. Instead of providing pixel-wise annotations, the expert delivers language feedback describing corrective actions on those regions. \textit{LINGUAL} then parses and executes these commands to generate a refined prediction mask $Y_{\text{AL}}$. Finally, $f_{\theta}$ is updated via backpropagation using supervised loss defined between $P_{\text{init}}$ and $Y_{\text{AL}}$ and any unsupervised loss on $P_{\text{init}}$, enabling progressive adaptation with minimal manual effort.

\subsection{Acquisition Function}
\label{AQF}

The objective of $\mathcal{A}$ is to identify and highlight regions in the segmentation map that exhibit the highest uncertainty with respect to the model predictions. At each active round, $\mathcal{A}$ ranks all candidate regions based on their uncertainty scores and selects a subset of regions $\{ R_i \}_{i=1}^{b}$ according to the predefined annotation budget per image. We employ several acquisition strategies from the literature, including Entropy~\cite{holub2008entropy}, RIPU~\cite{xie2022towards}, and ODES~\cite{islam2025odes} and so on to determine the most informative ROIs within the given budget constraint.


\subsection{Language Guidance}
\label{language_guidance}
After sampling ROIs, the expert provides language feedback there to align them with ground truth in those regions. We note that in a ROI, segmentation error generally occurs in the form of false positives (FP), false negatives (FN), internal holes within the foreground, disjoint fragmented structures, or irregular boundaries that require refinement through smoothing. In order to address FP and FN regions during segmentation correction, high-level natural language instructions—such as \textit{“Shrink”} for reducing FP and \textit{“Expand”} for filling in FN regions—can be used, supplemented by directional cues to guide the refinement process. In addition, multiple correction operations can be required for a single ROI. For example language-based feedback from an expert for particular ROI may be expressed as: \textit{“Expand the boundary at the top-right corner, remove the fragments at the bottom, and smooth the overall boundary."} A key question arises: \textit{How can these commands expressed in natural language effectively translated and executed in an automated way?} Motivated by  \cite{gupta2023visual} and the reasoning capacity of the Large Language Model \cite{gupta2023visual}, we decompose complex visual tasks based on natural language commands into interpretable and executable modules.

To facilitate modular and interpretable correction in medical image segmentation, we adopt a two-step process: (a) Program Generation (b) Program Execution. The program generation phase employs in-context learning to translate natural language instructions into a structured sequence of programs. Manually curated examples—consisting of language commands and their corresponding programs—are provided to GPT-3.5 \cite{openai2023gpt35turbo}, enabling it to generate similar program sequences in response to new instructions from the expert (Figure \ref{fig:visprog}(a)). 
The execution phase autonomously carries out these programs to perform the intended task through some pre-defined modules (Figure \ref{fig:visprog} b).

\begin{figure*}[htbp]
    \centering
    \includegraphics[width=0.97\linewidth]{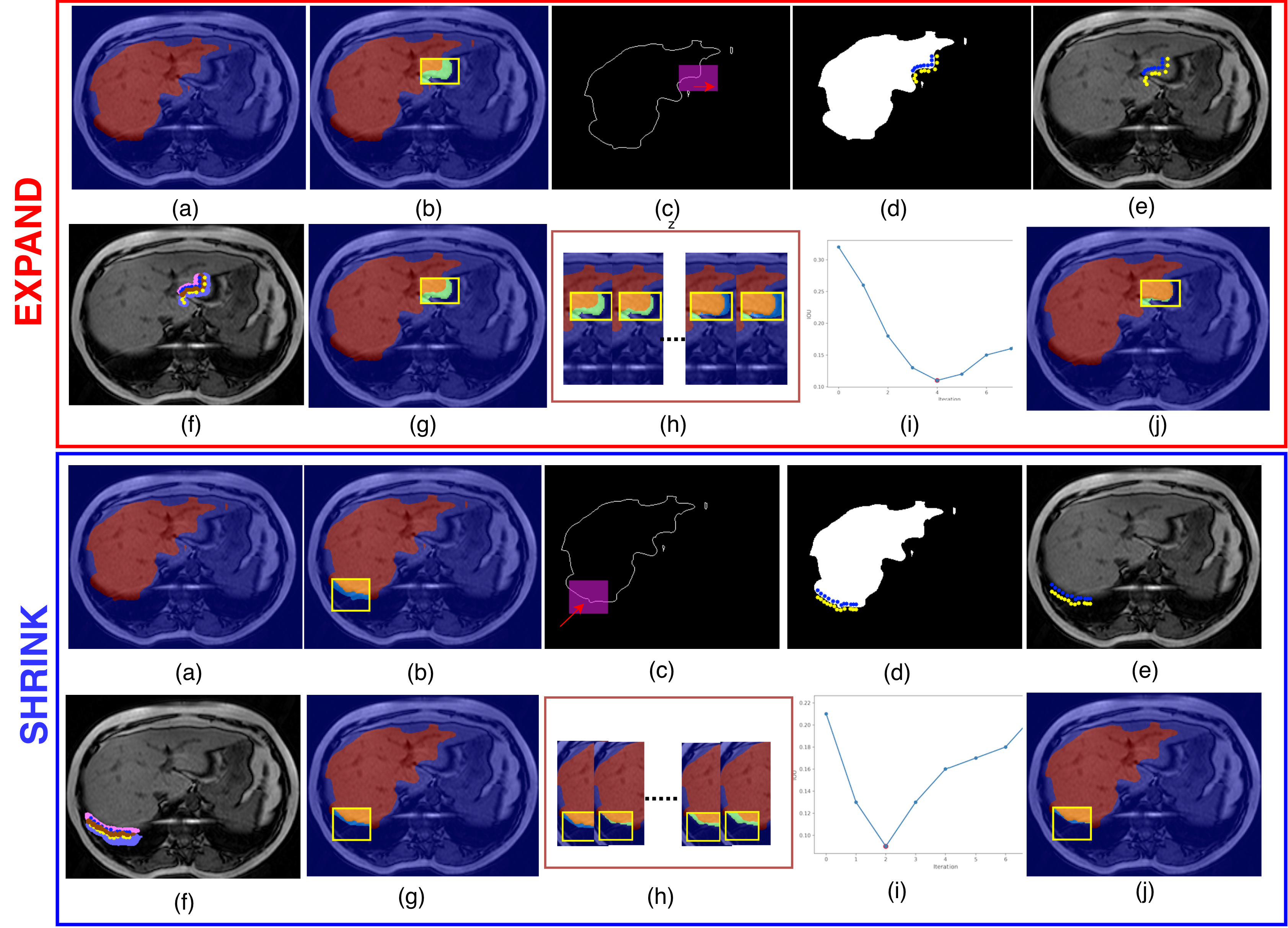}
    \caption{Illustration of each step involved in the \textit{EXPAND} and \textit{SHRINK} operations: The corresponding language commands are \textit{"EXPAND to Right"} and \textit{"SHRINK at Top-Right"}. The process is depicted as follows: (a) Initial prediction map (\textcolor{red}{red}) overlaid on the input image. (b) The yellow bounding box (ROI) obtained by AQF. Inside the ROI, true positive (\textcolor{orange}{orange}), false positive (\textcolor{blue}{blue}), and false negative regions (\textcolor{green}{green}) are shown.  
    (c) Boundary of prediction map along with the direction of operation shown by red arrow. (d) Sampled points from the inside (\( P_{\text{in}} \)) and outside (\( P_{\text{out}} \)) of boundary within ROI are shown as \textcolor{blue}{blue} and \textcolor{yellow}{yellow} dots, respectively. (e) \( P_{\text{in}} \) and \( P_{\text{out}} \) are also highlighted on the input image. (f) \( A_{\text{in}} \) (\textcolor{pink}{pink})  and \( A_{\text{out}} \) (\textcolor{myviolet}{violet}) are highlighted.\( A_{\text{in}} \cap \)
\( A_{\text{out}} \) shown by \textcolor{brown}{brown} color. (g) Updated prediction after the first iteration using Equations~\ref{expand_update} and~\ref{shrink_update}. (h) Results after multiple iterations. (i) Plot of \( \eta_{t} \) versus iteration number \( t \); the minimum point on the plot indicates the best output. (j) Final refined prediction corresponding to the optimal iteration, showing notable improvement over the initial state in (b).}
    \label{fig:expand_shrink}
\end{figure*}

\subsection{Execution Tools}
 In our segmentation refinement task, \textbf{EXPAND}, \textbf{SHRINK}, \textbf{REMOVE}, \textbf{FILL}, \textbf{SMOOTH}, \textbf{FOREGROUND} and \textbf{BACKGROUND} are implemented as pre-defined standalone modules. Among these modules, \textbf{EXPAND} and \textbf{SHRINK} are newly designed by us, while the remaining ones are adapted from the literature, primarily consisting of classical morphological operations.

\subsubsection{EXPAND and SHRINK Modules}
\label{expand_shrink}

The \textbf{EXPAND} and \textbf{SHRINK} modules pushes the prediction boundary in a specified direction by the expert  (e.g., \textit{"Expand to right", "Shrink at top-left"}) to improve segmentation quality. The \textbf{EXPAND} module enlarges the boundary to address under-segmentation, while the \textbf{SHRINK} module contracts the boundary inward to reduce over-segmented regions. We propose a novel iterative strategy for performing expansion and shrinkage operations guided by user-specified directions within a ROI. 

Our approach is inspired by the pixel clustering method described in \cite{zhao2024graco}, which operates on an image using a user-selected pixel and a specified granularity ($0 < \text{granularity} < 1$). Given a selected pixel and a granularity  input, the method identifies and clusters spatially adjacent pixels that share similar semantic features. The granularity allows to control the desired level of precision. For example, in medical imaging, a low granularity value yields fine-grained segmentation around the input point, isolating small, semantically consistent regions such as a tumor sub-region. In contrast, a high granularity value results in broader segmentation, capturing larger surrounding regions. We adopt a low granularity ($ \text{granularity} = 0.1 $) to enable precise, incremental refinements—analogous to a small step size in optimization. This minimizes the risk of overcorrection, whereas high granularity may cause coarse updates and introduce new errors, similar to instability from a high learning rate in optimization.

At first we highlight the sub-region of the prediction boundary based on expert-defined direction. We sample two sets of points, \( P_{\text{in}} \) and \( P_{\text{out}} \), each containing \( n \) points ( $n = s\%$ of highlighted boundary points, where $s$ is a hyper-parameter.), positioned near the inner and outer sides of the predicted segmentation boundary along the expert-defined direction, respectively. Let \( P_{\text{in}} = \{ p_{\text{in}}^{(i)} \}_{i=1}^n \) and \( P_{\text{out}} = \{ p_{\text{out}}^{(i)} \}_{i=1}^n \). Applying the clustering method \cite{zhao2024graco} to each point in \( P_{\text{in}} \) and \( P_{\text{out}} \), yields clustered regions \( A_{\text{in}}^{(i)} \) and \( A_{\text{out}}^{(i)} \), respectively. The total regions captured by \( P_{\text{in}} \) and \( P_{\text{out}} \) are defined as:

\[
A_{\text{in}} = \bigcup_{i=1}^{n} A_{\text{in}}^{(i)}, \quad 
A_{\text{out}} = \bigcup_{i=1}^{n} A_{\text{out}}^{(i)}.
\]
At iteration \( t \), \( A_{\text{in}} \) and \( A_{\text{out}} \) are denoted as \( A_{\text{in}}(t) \) and \( A_{\text{out}}(t) \), respectively. The predicted segmentation mask \( P_t \) is updated as:

\textbf{EXPAND:}
\begin{equation}
\label{expand_update}
P_{t+1} = P_t \cup \left( A_{\text{in}}{(t)} \cap A_{\text{out}}{(t)} \right)
\end{equation}

\textbf{SHRINK:}
\begin{equation}
\label{shrink_update}
P_{t+1} = P_t \setminus \left( A_{\text{in}}{(t)} \cap A_{\text{out}}{(t)} \right)
\end{equation}

We note that when \( P_{\text{in}} \) and \( P_{\text{out}} \) are sampled from the same semantic region, the regions \( A_{\text{in}} \) and \( A_{\text{out}} \) may significantly overlap. In contrast, if the points are sampled from different semantic regions (e.g., opposite sides of a true boundary), the overlap between \( A_{\text{in}} \) and \( A_{\text{out}} \) is minimal.
  This is because, near the true boundary, pixels on opposite sides tend to exhibit high semantic dissimilarity. Consequently, the clustering mechanism, which is guided by semantic coherence, inherently restricts the inclusion of pixels from neighboring semantic regions when a seed point is placed near the boundary. Therefore, in an iterative correction framework, the intersection \( A_{\text{in}} \cap A_{\text{out}} \) is minimized when the predicted boundary aligns closely with the true boundary. To account for variations in the number of sampled points \( n \) across iterations, we compute a normalized overlap metric \( \eta \) for iteration $t$, defined as:
\begin{equation}
    \label{eta}
    \eta_{t} = \frac{|A_{\text{in} }(t) \cap A_{\text{out}} (t)|}{|A_{\text{in}}(t) \cup A_{\text{out}}(t)|}
\end{equation}
which serves as a scale-invariant measure of the overlap between regions sampled from opposite sides of the predicted boundary. A lower value of \( \eta \) indicates higher semantic separation between $P_{in}$ and $P_{out}$ which indirectly indicates the corrected segmentation boundary is closer to the true boundary. Our proposed method for \textbf{EXPAND} and \textbf{SHRINK} is shown in Figure \ref{fig:expand_shrink}. The algorithm is also provided in the supplementary (algorithm 1).

\subsubsection{Other Modules}

In an erroneous segmentation maps, it is common to observe  small disjoint fragments that do not belong to the true foreground, small holes within the predicted foreground regions, and rough or jagged external boundaries. Also there can be regions sampled by the $\mathcal{A}$ that is entirely foreground or background.To address these issues, we incorporate classical morphological operations in a modular fashion to perform localized corrections.

\textbf{REMOVE:}  
To eliminate small disconnected fragments from a ROI of the predicted mask $P$, we perform connected component analysis using 8-connectivity. Components with area less than a threshold are removed, retaining only spatially significant structures.This morphological filtering step is widely used in post-processing for segmentation refinement \citep{shapiro2001connected}.

\textbf{FILL:}  
To fill up small holes in a binary mask, we incorporate morphological closing \citep{serra1982image}. This involves applying a dilation followed by an erosion.

\textbf{SMOOTH:}  To reduce jaggedness along the external boundary of the predicted mask, we apply \textit{Gaussian smoothing}~\cite{gonzalez2007digital} followed by thresholding.

\textbf{DIRECTION: }This module extracts the direction keyword from the annotator's natural language commands to perform the correction operation (e.g., \textit{“shrink from the right”}, or \textit{“expand at bottom”} or \textit{“smooth the right border}). The direction is defined relative to the interior of the ROI. There are 8 directions which are considered: top, bottom, left, right, top-left, top-right, bottom-left and bottom-right. These directional commands can be used with any of EXPAND, SHRINK, REMOVE, FILL or SMOOTH operation. If no specific direction is provided for an operation, it is applied to the entire ROI. For instance, the instruction \textit{"Smooth the overall boundary"} implies smoothing the entire boundary within the ROI.

\textbf{FOREGROUND:} This module assigns the entire region as foreground.

\textbf{BACKGROUND:} This module assigns the entire region as background.

\textbf{RESULT: }This module collects and summarizes the final corrected segmentation output after all visual program steps have been executed.

\begin{table*}[t]
\centering
\footnotesize
\caption{Performance comparison on IP$\rightarrow$OOP and BUSI$\rightarrow$UDIAT. All numbers represent Dice scores (\%). GT ROI means ROI is annotated using GT, Poly menas ROI is annotated using polygonal approximation. We also report the boost of performance compared to source model in \textit{LINGUAL}.} 
\label{adaptation_methods}
\begin{tabular*}{0.95\textwidth}
{@{\extracolsep{\fill}}l | c c c c c c c | c c}
\toprule
& & \multicolumn{6}{c|}{\textbf{IP $\rightarrow$ OOP}} & \multicolumn{2}{c}{\textbf{BUSI $\rightarrow$ UDIAT}} \\
\midrule
\textbf{Method}& \textbf{ROI type} &\textbf{Budget} & \textbf{Liver} & \textbf{L.Kidney} & \textbf{R.Kidney} & \textbf{Spleen} & \textbf{Average} & \textbf{Budget} & \textbf{Tumor} \\
\midrule
Source Model & N/A &N/A & 87.36 & 47.22 & 36.60 & 44.11 & 53.82 & N/A & 66.43 \\

Fully Supervised &N/A &100\% & 93.65 & 85.61  & 87.06 & 83.45 & 87.44 & 100\% & 81.19 \\

Random Selection &Patch &5\% & 88.84 & 60.73 & 68.24 & 55.54 & 68.34 & 5\% & 71.83 \\
\hdashline[0.5pt/2pt]

Entropy \cite{holub2008entropy} (GT ROI ) &Patch & 5\% & 92.45 & 87.23 & 86.18 & 77.1 & 85.74 & 5\% & 79.73 \\
Entropy (Poly) &Patch & 5\% & 92.02 & 84.72 & 84.54 & 78.43 & 84.93 & 5\% & 78.82 \\
Entropy + LINGUAL &Patch &5\% & 92.35 & 80.29 & 83.94 & 77.92 & 83.63 (+29.81 $\%$) & 5\% & 78.32 (+11.89 $\%$)\\

\hdashline[0.5pt/2pt]

RIPU \cite{xie2022towards} (GT ROI) &Patch &5\%& 92.05&  86.35&  87.26&	79.95&	86.40 & 5\% & 79.44 \\

RIPU (Poly) &Patch &5\%& 90.88 & 84.96 & 85.88 & 77.47 & 84.8 & 5\% & 78.41 \\
RIPU + LINGUAL &Patch &5\% & 92.56&	82.01&	85.77&	77.02&	84.34 (+30.52 $\%$)  & 5\% & 78.61 (+12.18 $\%$)\\
\hdashline[0.5pt/2pt]




RIPU ROI+MEDSAM \cite{ma2024segment}&   Patch&  5\%& 90.43&	84.63&	84.93&	68.05&	82.01&  5\%& 77.97 \\
RIPU ROI+LINGUAL &Patch     &5\% & 92.56&	82.01&	85.77&	77.02&	\textbf{84.34} (+30.52 $\%$) & 5\% & 78.61(+12.18 $\%$) \\
\midrule





ASAL \cite{kim2023adaptive} &SP  &25 SP & 91.68 & 79.69 & 78.89 & 69.66 & 79.98 & 25 SP & 77.11 \\
ASAL +LINGUAL &SP    &  25 SP&   91.27 & 81.88 & 81.19 & 73.14 & \textbf{81.87} (+28.05 $\%$) &  25SP & \textbf{78.28} (+11.85 $\%$) \\
\hdashline[0.5pt/2pt]




\hdashline[0.5pt/2pt]

\bottomrule
\end{tabular*}
\end{table*}

\subsection{Language Prompt Generation Strategy}
We note that no medical experts were involved in our experiments. Instead, we used publicly available datasets and treated the provided ground truth (GT) annotations as a proxy for expert feedback. Following the active learning protocol, we assumed that the GT is accessible only within the ROIs selected by $\mathcal{A}$, while all other regions remain untouched. By systematically comparing the prediction ($P_{init}$) and GT inside the ROI we first identify which tools are necessary to perform the refinement procedure. After identifying tools and using GPT-3.5 \cite{openai2023gpt35turbo}, we generate language-like commands in a reproducible manner. For example:
\textit{How to identify the ROI requires EXPAND/SHRINK operation?}

The decision process consists of two stages: identifying the required operation and estimating its direction. To determine whether an ROI requires an \textbf{EXPAND} or \textbf{SHRINK} update, we compare the prediction and ground-truth masks inside the ROI and compute the sets of False Negatives (FN) and False Positives (FP). If FN pixels dominate, the model has under-segmented the structure, indicating the need to push the boundary outward via an \textbf{EXPAND} operation; if FP pixels dominate, the model has over-segmented the region, requiring a \textbf{SHRINK} operation to contract the boundary. After identifying the correct operation, we determine the direction by first extracting the prediction boundary within the ROI and computing its centroid. For an EXPAND operation, we also compute the centroid of the FN region and draw a vector from the prediction-boundary centroid to the FN centroid; this vector encodes the direction along which the boundary should expand. Similarly, for a SHRINK operation, the direction vector is obtained by connecting the prediction-boundary centroid to the centroid of the FP region. The angle of this vector relative to the coordinate axes is then used to map the operation to one of the eight predefined directional commands (e.g., top, bottom-right, left, etc.). After determining the operation type and direction we utilize GPT 3.5 to produce expert-like language feedback. We provide the language prompts used to generate expert-like feedback, in the supplementary material.

\section{Experiments and Results}
In this section we shall present about the dataset, different experimental settings and the results.

\subsection{Datasets and Experimental Setup}
 

We evaluate our method in an \textbf{Active Domain Adaptation (ADA)} setting across two imaging modalities:  

\textbf{(a) MRI:} Multi-organ segmentation (liver, left kidney, right kidney, spleen) using the CHAOS T1-Dual dataset, adapting from \textbf{In-Phase (IP)} to \textbf{Out-of-Phase (OOP)} images~\cite{kavur2021chaos}.  

\textbf{(b) Ultrasound:} Breast tumor segmentation, adapting from \textbf{BUSI}~\cite{al2020dataset} to \textbf{UDIAT} (B$\rightarrow$U)~\cite{byra2020breast}.  

In both settings, the source datasets (IP and BUSI) are used to train an initial source model $f_\theta$, which is subsequently adapted to the corresponding target domains (OOP and UDIAT). According to \cite{byra2020breast,thomas2023bus,hu2021fully, islam2025odes}, these adaptation scenarios exhibit domain shift. We utilize Deeplabv3+ architecture \cite{chen2018encoder} as our segmentation model. We train the model to segment two-dimensional images. For MRI data every slice is considered as the input image. The budget is set as $5\%$ regions/image for patch based AL and 25 superpixels (SP)/image for SP-based AL. We choose $21 \times 21$ as our base ROI size. The supplementary document provides additional details on this.

\subsection{Performance Comparison}
\label{perforamance_comparison}
We compare \textit{LINGUAL} under patch-level and superpixel (SP)-level AL. Since no human experts are involved in our experiments, we use the ground truth (GT) within the sampled ROIs as a proxy for expert annotations. However, medical images often exhibit highly complex boundaries. Experts generally annotate these regions using polygonal delineation, but even for experienced annotators, achieving pixel-level precision is challenging. The intricate curvature along object boundaries inevitably introduces a margin of error in polygon-based annotations. To simulate this effect, we employ a curvature-adaptive boundary sampling algorithm \cite{douglas1973algorithms} where boundary contour vertices are adaptively sampled based on local curvature magnitude to ensure full boundary coverage with minimal points. Using these vertices we can get the polygonal approximation which is used for labeling.




In Table \ref{adaptation_methods}, we compare \textit{LINGUAL} with state-of-the-art (SOTA) AL methods. For patch-based SOTA approaches, performance is evaluated using (i) GT annotations within the sampled ROIs, and (ii) simplified polygonal approximations of the GT. We observe a small performance gap between both cases and \textit{LINGUAL}, which is expected because (i) directly uses GT sub-regions and (ii) uses its simplified approximation. Despite relying solely on high-level language feedback without any manual annotation, \textit{LINGUAL} achieves performance comparable to (ii), demonstrating the effectiveness of our approach. 




\textit{LINGUAL} outperforms SP-based AL methods (on average 3.13 $\%$ and 1.78 $\%$ each adaptation ) under the same annotation budget. SP based AL can not always encounter ambiguous and blurry boundaries which lead to sub-optimal SP generation, introducing labeling noise that degrades model performance. In contrast, \textit{LINGUAL} employs an iterative boundary-refinement strategy with fine-grained corrections, enabling more precise boundary identification and superior segmentation quality. 

\subsection{Discussion on Manual Effort}
\label{sec_effort}
 We compare the manual effort required for conventional polygonal delineation with that of the proposed \textit{LINGUAL} framework.  For simplicity, we assume manual effort is equivalent to manual annotation time. In manual delineation, experts must carefully place vertices along the object boundary, sequentially forming a closed polygon that represents the foreground. The annotation effort is therefore proportional to both the total boundary length and its curvature complexity. In order to simulate polygonal delineation we employ \cite{douglas1973algorithms} as discussed in Sec. \ref{perforamance_comparison} . According to authors in \cite{benenson2019large}, manual polygonal annotation typically requires 5.55 secs per vertex. From these information, we estimate the time for polygon-based delineation. For \textit{LINGUAL}, manual effort is primarily verbal. We assume the presence of an automatic speech-to-text module~\cite{trivedi2018speech} that converts the expert’s spoken feedback into text, after which the remaining workflow proceeds autonomously.

 To quantify human effort, we measure the number of words spoken during expert feedback. According to \cite{venkatagiri1999clinical}, the average speech rate is 130 words per minute during describing an image. Using these information we estimate the time required for language feedback. Table \ref{manual_effort} shows that \textit{LINGUAL} reduces annotation time 83.4 $\%$ and 79.9 $\%$ compared to polygonal delineation on two adaptations. 
 
 In Table \ref{manual_effort} we also demonstrated manual effort estimation during scaling of each ROI area size (number of ROI remains the same). It shows doubling each of ROI size increases the annotation time of polygonal delineation by 36.98 $\%$ and 52.91 $\%$. In contrast, annotation time increases very little (1.75 $\%$ and 1.16 $\%$) in \textit{LINGUAL} despite scaling the ROI size. Specifically, in \textit{LINGUAL}, the manual annotation time or manual effort scales with the number of ROIs, as the expert provides distinct feedback for each region. However, it remains largely independent of the ROI size.

\begin{table}[!h]
\centering
\scriptsize
\caption{Annotation time comparison between polygonal delineation and \textit{LINGUAL} across ROI area scales.}
\begin{tabular}{l l c c c c c}
\toprule

\textbf{Adapt.} & \textbf{Scale} & \multicolumn{2}{c}{\textbf{Poly. Delin.}} & \multicolumn{2}{c}{\textbf{\textit{LINGUAL}}} & \textbf{{$\Delta_{\text{time}}$}} \\
\cmidrule(r){3-4} \cmidrule(r){5-6}
 &  & \#Vertices & Time(hr) & \#Words & Time(hr) &  \\
\midrule
B$\rightarrow$U & 1$\times$ & 8,943 & 13.79 & 17,880 & 2.29 & -83.4 $\%$ \\
IP$\rightarrow$OOP & 1$\times$ & 27,826 & 42.90 & 67,186 & 8.61 & -79.9 $\%$\\
\midrule
B$\rightarrow$U & 2$\times$ & 12,318 & 18.99 & 18,211 & 2.33 & -87.7 $\%$ \\
IP$\rightarrow$OOP & 2$\times$ & 42,550 & 65.60 & 67,988 & 8.71 & -86.8 $\%$ \\
\bottomrule
\end{tabular}
\label{manual_effort}
\end{table}

\subsection{Ablation Study}

\begin{table}[!h]
\centering
\scriptsize
\caption{Effect of ROI size (area) scaling on segmentation performance (DSC). Number of ROI per round shown by ($N_{\text{ROI}}$).}
\label{scaling}
\begin{tabular}{c c c c c c}
\toprule
\textbf{Case} & $\text{AREA}_{\text{ROI}} \times$ & $N_{\text{ROI}} \times$ & \textbf{Man. Eff.} $\times$ & \textbf{B$\rightarrow$U} & \textbf{IP$\rightarrow$OOP} \\
\midrule
1 & 1$\times$ & 1 $\times$ & 1$\times$ & 72.62 & 65.30 \\
2 & 2$\times$ & 0.5 $\times$ & 0.5$\times$ & 69.29 & 62.34 \\
3 & 2$\times$ & 1 $\times$ & 1$\times$ & 74.31 & 70.81 \\
\bottomrule
\end{tabular}
\end{table}


\noindent \textbf{\textit{Effects of ROI Size Scaling.}}We evaluate the impact of ROI size scaling under two conditions: (i) keeping the total annotated area constant and (ii) keeping the number of ROIs constant. Table~\ref{scaling} summarizes the results. Case~1 serves as the baseline, with $N_{\text{ROI}}=2$ and three active rounds. In Case~2, the ROI size is doubled while the number of ROI is reduced proportionally to maintain the same total annotated area. This leads to a performance decline, as fewer ROIs reduce the likelihood of sampling the most informative regions.

In Case~3, the ROI size is increased but the number of ROI remains unchanged, resulting in a larger total annotated and improved segmentation performance. Manual effort remains comparable to Case~1 since, as discussed in Section~\ref{sec_effort}, annotation time in \textit{LINGUAL} depends on the number of ROIs rather than their individual area. Consequently, \textit{LINGUAL} enhances the scalability of AL without increasing human effort.






\section{Conclusion}
We introduced \textbf{\textit{LINGUAL}}, the first language-guided AL framework for medical image segmentation. Instead of dense, sequential polygonal delineation, \textit{LINGUAL} leverages high-level language feedback that is automatically translated into executable refinement programs, substantially reducing human effort. The framework achieves segmentation accuracy close to patch-based AL while reducing annotation time by approximately 80 $\%$, and further outperforms superpixel-based AL. This paradigm highlights the potential of language as an intuitive supervisory signal, paving the way for more efficient and interactive human--AI collaboration in medical image annotation.



{
    \bibliographystyle{ieeenat_fullname}
    \bibliography{main}
}


\end{document}